\documentclass[sigconf]{acmart}

\usepackage[utf8]{inputenc} 
\usepackage[T1]{fontenc}    
\usepackage{hyperref}       
\usepackage{url}            
\usepackage{booktabs}       
\usepackage{amsfonts}       
\usepackage{nicefrac}       
\usepackage{microtype}      
\usepackage{xcolor}         
\usepackage{amsmath}
\usepackage{amsthm}
\usepackage{graphicx}
\usepackage{multirow}
\usepackage{textcomp}
\usepackage{subfig}

\usepackage{wrapfig}
\usepackage{algorithm}
\usepackage{algorithmic}

\AtBeginDocument{%
  }

\copyrightyear{2023}
\acmYear{2023}
\setcopyright{acmlicensed}\acmConference[KDD '23]{Proceedings of the 29th ACM SIGKDD Conference on Knowledge Discovery and Data Mining}{August 6--10, 2023}{Long Beach, CA, USA}
\acmBooktitle{Proceedings of the 29th ACM SIGKDD Conference on Knowledge Discovery and Data Mining (KDD '23), August 6--10, 2023, Long Beach, CA, USA}
\acmPrice{15.00}
\acmDOI{10.1145/3580305.3599385}
\acmISBN{979-8-4007-0103-0/23/08}

\newcommand{\vx}{\mathbf{x}}
\newcommand{\vy}{\mathbf{y}}

\newcommand{\reffig}[1]{Figure. \ref{#1}}

\newcommand{\reftab}[1]{Table. (\ref{#1})}
\newcommand{\refsec}[1]{Section. \ref{#1}}

\begin{document}

\title{IDToolkit: A Toolkit for Benchmarking and Developing Inverse Design Algorithms in Nanophotonics}

\author{Jia-Qi Yang}
\email{yangjq@lamda.nju.edu.cn}
\affiliation{%
  \institution{State Key Laboratory for Novel Software Technology\\Nanjing University}
  \city{Nanjing}
  \country{China}
  \postcode{210023}
}

\author{Yucheng Xu}
\email{yucheng.xu@smail.nju.edu.cn}
\affiliation{%
  \institution{Research Institute of Superconductor Electronics (RISE), School of Electronic Science and Engineering, Nanjing University}
  \city{Nanjing}
  \country{China}
  \postcode{210023}
}

\author{Jia-Lei Shen}
\email{191300043@smail.nju.edu.cn}
\affiliation{%
  \institution{State Key Laboratory for Novel Software Technology\\Nanjing University}
  \city{Nanjing}
  \country{China}
  \postcode{210023}
}

\author{Kebin Fan}
\email{kebin.fan@nju.edu.cn}
\affiliation{%
  \institution{Research Institute of Superconductor Electronics (RISE), School of Electronic Science and Engineering, Nanjing University}
  \city{Nanjing}
  \country{China}
  \postcode{210023}
}

\author{De-Chuan Zhan}
\authornote{Corresponding authors.}
\email{zhandc@nju.edu.cn}
\affiliation{%
  \institution{State Key Laboratory for Novel Software Technology\\Nanjing University}
  \city{Nanjing}
  \country{China}
  \postcode{210023}
}

\author{Yang Yang}
\email{yyang@njust.edu.cn}
\authornotemark[1]
\affiliation{%
  \institution{Nanjing University of Science and Technology}
  \city{Nanjing}
  \country{China}
  \postcode{210023}
}

\begin{abstract}
  Aiding humans with scientific designs is one of the most exciting
  of artificial intelligence (AI) and machine learning (ML),
  due to their potential for the discovery of new drugs,
  design of new materials and chemical compounds, etc.
  However, scientific design typically requires complex domain knowledge
  that is not familiar to AI researchers.
  Further, scientific studies involve professional skills to perform
  experiments and evaluations.
  These obstacles prevent AI researchers from developing specialized
  methods for scientific designs.
  To take a step towards easy-to-understand and reproducible
  research of scientific design,
  we propose a benchmark for the inverse design of nanophotonic devices, 
  which can be verified computationally and accurately.
  Specifically, we implemented three different nanophotonic design problems,
  namely a radiative cooler,
  a selective emitter for thermophotovoltaics,
  and structural color filters,
  all of which are different in design parameter spaces, complexity, and design targets. 
  The benchmark environments are implemented with an open-source simulator.
  We further implemented 10 different inverse design algorithms
  and compared them in a reproducible and fair framework.
  The results revealed the strengths and weaknesses of existing methods,
  which shed light on several future directions for developing more efficient inverse design algorithms. Our benchmark can also serve as the starting point for more challenging scientific design problems.
  The code of IDToolkit is available at \url{https://github.com/ThyrixYang/IDToolkit}.
\end{abstract}

\ccsdesc[500]{Applied computing~Physical sciences and engineering}


\keywords{Benchmark, Inverse Design, Datasets}


\maketitle

\section{Introduction}

Machine learning for scientific discovery,
with the goal of establishing data-driven methods to advance
the research and developments,
has become an intriguing area attracting a lot of researchers
in recent years. \cite{ml4sci_1,ml4sci_2,ml4sci_3,ml4sci_4}.
For example, machine-learning techniques have been successfully
applied in biology \cite{alphafold}, chemistry \cite{chemical_design},
physics\cite{ml_physics}, nanophotonics\cite{nf_id_2023} etc.
In the majority of scientific design problems, 
the performance of a design often necessitates physical experiments to validate its efficacy. 
This greatly hinders the development and validation of inverse design algorithms. 
On the contrary, the performance of nanophotonic devices can be verified through computational simulations accurately, 
enabling a significant acceleration in the verification process and a reduction in associated costs.

\begin{figure*}[h]
  \includegraphics[width=\textwidth]{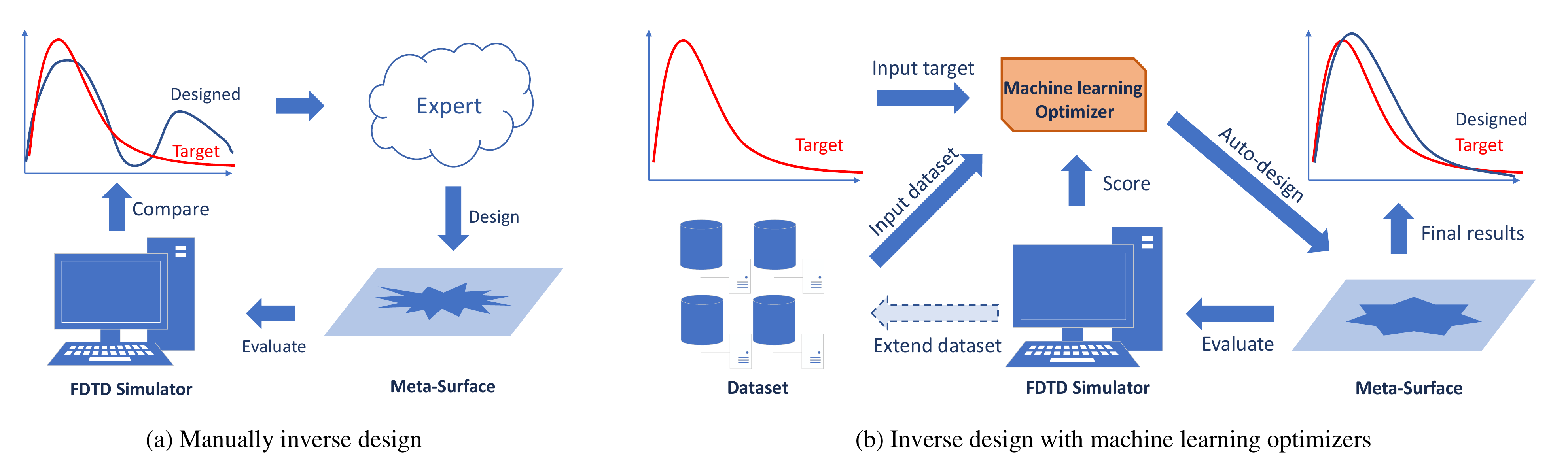}
  \caption{The procedure of manually inverse design and
    our framework of inverse design based on machine learning.
    \label{fig:intro}
  }
\end{figure*}
In nanophotonics, inverse design refers to approaches that
aim to discover optical structures with fashioned optical properties.
Generally, a "forward" procedure means the mapping from the given
geometry of a nanophotonic device to the target electromagnetic response.
Numerical methods can obtain this mapping process.
While an "inverse" procedure indicates a backward process from the
target optical response to predict some nanophotonic structures with
properties close to the target response.
A traditional inverse design process of nanophotonic structures highly
rely on the experiments of human experts as depicted in \reffig{fig:intro} (a).
But this method is hard to design complex nanophotonic structures with
a lot of design parameters.
Various computational inverse-design methods have been proposed
to improve the design efficiency for high-performance nanophotonic
devices,
such as genetic algorithms~\cite{baumert1997femtosecond},
and variations of the adjoint method~\cite{LalauKeraly2013, sell2017large, xiao2016diffractive, Hughes2018, Mansouree2021}.
Among them, the adjoint method,
which requires explicit formulation of the underline Maxwell's equation,
is one of the most widely adopted methods in commercial simulation software.
However, this method cannot tackle discrete parameters and require
a large number of forward simulations, as well as some treatments to jump out of
the local minima.\cite{Campbell2019}
Thus, data-driven, problem-agnostic inverse design methods based on
machine learning approaches are of particular interest in nanophotonic research
in recent years.

However, some prominent obstacles
prevent machine learning researchers from developing inverse design algorithms.
First, it is not straightforward for computer scientists to understand
the specific problems in other scientific areas.
The critical issue lies in the formulation of an inverse design problem
into a machine learning problem due to the required domain knowledge.
Secondly, it is challenging to evaluate the performance of the design. 
For example, a new design might need to be simulated or
produced to measure performance.
However, it's nearly impossible for machine-learning researchers to
evaluate by themselves because the evaluation may need professional software,
instruments, and skills to operate.
These obstacles prevent the machine-learning community from
developing practical algorithms independently.
Thirdly, most nanophotonic studies did not open-source the implementation
of the simulation model and the inverse design algorithms,
which further exacerbated the difficulties of reproducing and
comparing inverse design algorithms.

Successful benchmarks can directly impel the development of machine learning algorithms.
Machine-learning-based data-driven approaches rely on
established datasets and benchmarks to develop and evaluate.
To name a few, Imagenet\cite{datasets_imagenet} for image classification,
COCO\cite{datasets_coco} for object segmentation,
OpenAI Gym\cite{datasets_opoenaigym} and
StarCraft\cite{datasets_starcraft} for reinforcement learning,
Movielens\cite{datasets_movielens} for recommender systems, etc.
There are a few inverse design benchmarks that rely on
substitute models to evaluate the performance\cite{benchmark_na,benchmark_na2}.
However, the simulation error of the inverse-designed parameters
turned out to be significantly larger
than predicted by the substitute model\cite{benchmark_na_j},
which indicates that direct evaluation with a simulator is
necessary for developing and evaluating accurate inverse design algorithms.
As far as we know, there lacks an inverse design benchmark
with direct interaction with physical simulators.

In this work, we introduce a benchmark for inverse design,
which consists of three different nanophotonic designs from practical applications.
We implemented the simulation models of all three problems with MEEP\cite{meep},
which is open-source and scalable on Linux clusters.
We implement 10 different inverse design algorithms,
and conducted extensive experiments.
The experimental results indicate that
the performances are related to the properties of specific problems and algorithms.
For example, a tree-based optimization algorithm performs better
in design problems with discrete parameters;
gradient descent method performs better in scenarios
with a large amount of training data and continuous parameters.
In addition, we point out several promising research directions
based on the experimental results and our experience.
Our benchmark can also serve as the starting point for other challenging
scientific design problems, such as chemical design\cite{chemical_design}.
The algorithms and interfaces implemented in the benchmark
can also be adapted to different design problems.
Our contribution can be summarized as follows:
\begin{itemize}
  \item We propose three distinct inverse design problems selected from practical applications,
        which have different search spaces and various difficulties.
        These problems can be used to develop and evaluate inverse design algorithms
        in a unified framework.
  \item We implemented and compared 10 different inverse design methods that cover
        most of the existing approaches in inverse design literature.
        We conduct extensive experiments to compare the algorithms from different perspectives,
        which reveals the strength and weaknesses of different algorithms.
  \item Based on our experimental results and our analysis
        of the inverse design problems and algorithms,
        we point out several future directions for developing
        efficient inverse design algorithms with our benchmark.
\end{itemize}

\section{Background}

In inverse design problems, we are supposed to have a forward simulation interface,
which takes the design parameters $\vx$ as input and output
a vector $\vy$ that corresponds to some concerned properties
that are related to the performance.
Generally, a forward simulation procedure can be
denoted as $\mathcal{F}(\vx)=\vy$, where $\vx=\{x_0, ..., x_{n-1}\}$
denotes $n$ design parameters,
$\vy=\{y_0,...,y_{m-1}\}$ denotes $m$ design targets.
The design parameters $\vx_i$ may be continuous or discrete,
for example, in the color filter problem, we need to design
the material and thickness of a specific layer,
where the material is a discrete design parameter that can be chosen within a pool,
and the thickness is a continuous design parameter.
The design targets $\vy$ are continuous in our problems.

The goal of inverse design is to find the best design parameter $x^*$ that
\begin{equation}
  \begin{aligned}
    \vx^* = \min_{\vx} \quad & \mathcal{L}(\mathcal{F}(\vx), \vy_{\text{target}}) \\
    \textrm{s.t.} \quad      & \vx \in \mathcal{D}
  \end{aligned}
\end{equation}
where $\mathcal{D}$ is the feasible design parameter space.
$\mathcal{L}(\vy, \vy^\prime)$ is a function that measures the difference between
$\vy$ and $\vy^\prime$.
The search space $\mathcal{D}$ of design parameters can be complex.
For example, in the TPV inverse design problem, the feasible range of
$\vx_3$ is determined by the value of $\vx_0$.
Typically, perfect solution with $\mathcal{F}(\vx)=\vy_{target}$ does not
exist, which further increases the difficulty.

\section{The benchmark}

In this section, 
we provide a brief overview of the composition of input variable $\vx$ and output variable $\vy$ 
for the three mentioned problems. 
A detailed description of the physical problems is presented in the appendix.

\subsection{Multi-layer Optical Thin Films (MOTFs)}

\begin{figure}[h]
  \includegraphics[width=\linewidth]{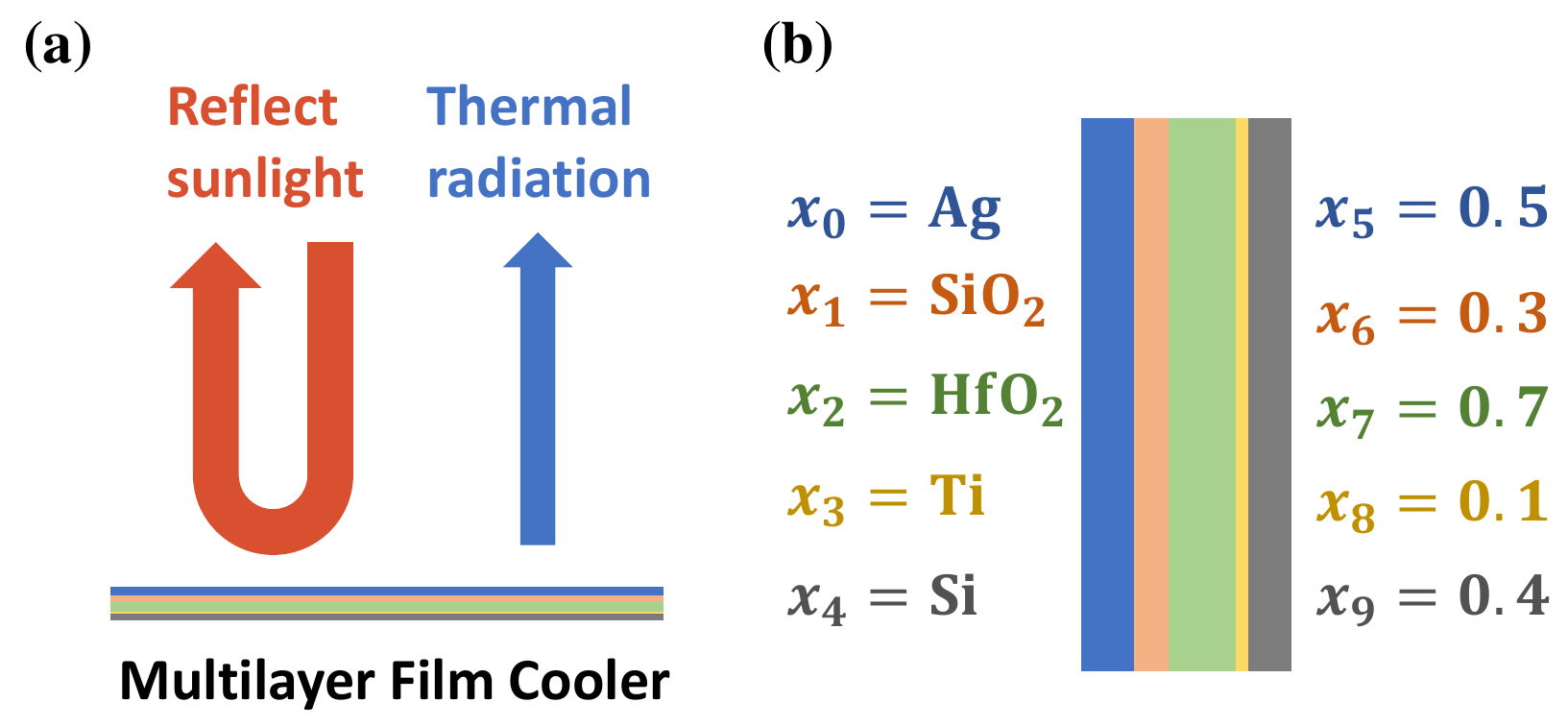}
  \caption{Depiction of MOTFs:
    (a) How a MOTFs works.
    (b) The design parameters of a MOTFs with 5 layers.
    \label{multilayer_film_intro}}
\end{figure}


Multilayer optical thin films (MOTFs) are simple structures with many layers of 
optical films stacked together. 
They have been widely used in many optical applications to achieve specific optical properties, 
such as antireflective coating (ARC), distributed Bragg reflector (DBR), energy harvesting, 
and radiative cooling.
To optimize an MOTF cooler with ideal properties, we implement a simulator with 7 different dispersive materials. 
We set the layer number to 10, which provides a fairly large search space and enough flexibility (\citet{multi_layer_nature14} proposes a 10 layer design). 
The layer number can be easily increased if necessary and less layer number can be obtained through setting some layer thicknesses to 0. 
In addition, the MOTF cooler design problem has both discrete (material type) and continuous (layer thickness) variables, 
and the effective number of variables is not fixed. 
The design parameters of a multilayer film cooler is depicted in \reffig{multilayer_film_intro} (b). 
Specifically, assuming there $k$ different layers, the design parameter $\vx_0, ..., \vx_{k-1}$ determine the the material used in the 1st to the kth layer respectively; 
the design parameter $\vx_k, ..., \vx_{2k-1}$ define the thicknesses of the 1st to kth layer, respectively. 
The material can be set to 7 different categorical values $\vx_i \in \{ZnO, AlN, Al_2 O_3, MgF_2, SiO_2, TiO_2, SiC\}, 0<i<k$. 
The layer thicknesses can be set to $\vx_i \in [0,1], k\le i < 2k$.
The response $\vy$ is a 2001-dimensional real-valued vector.

\subsection{Thermophotovoltaics (TPV)}

\begin{figure}[h]
  \includegraphics[width=\linewidth]{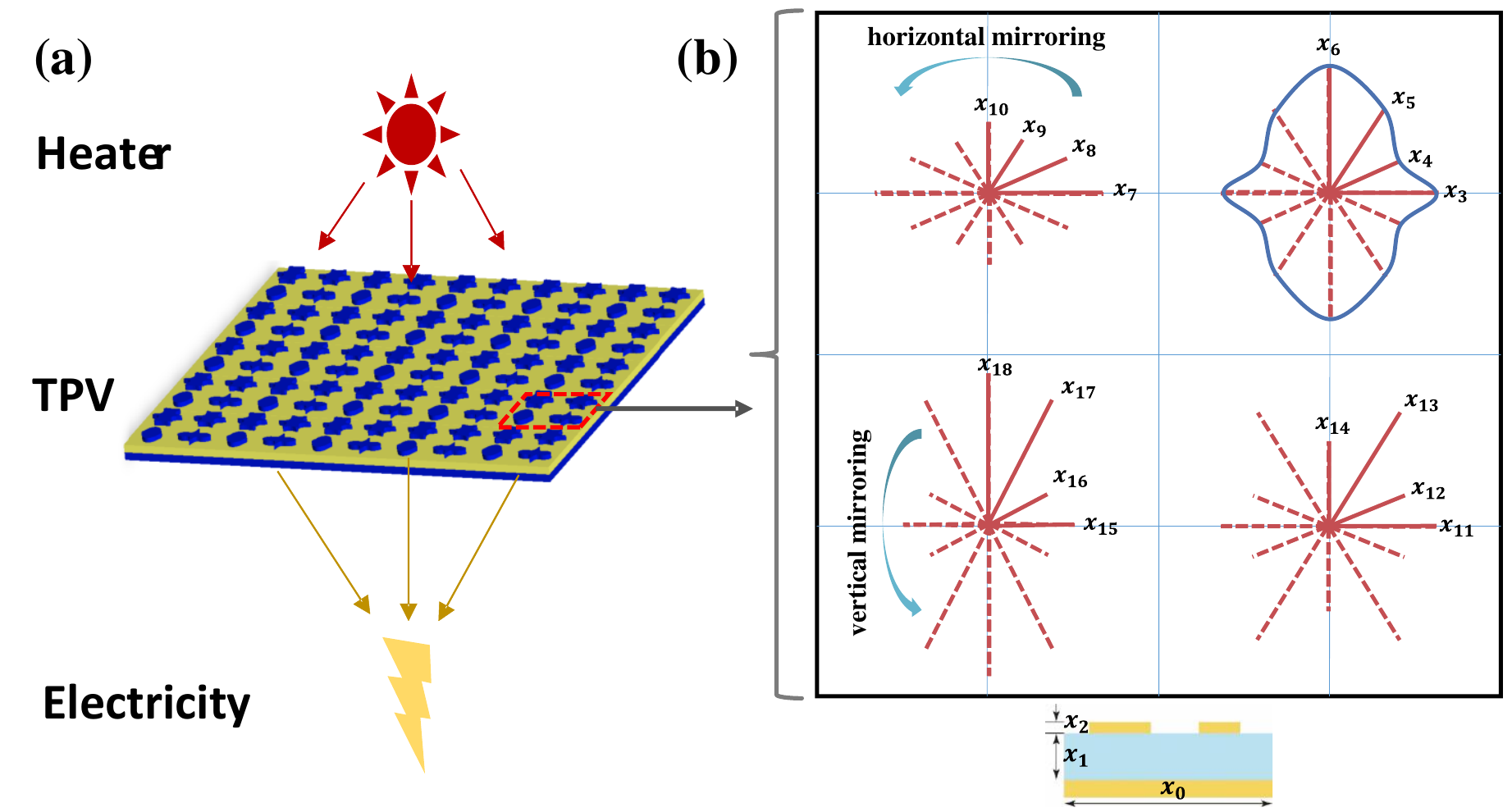}
  \caption{Depiction of the TPV problem:
    (a) How TPV energy conversion works.
    (b) The design parameters of TPV.\label{tpv_intro}}
\end{figure}


One of the promising applications of metamaterials is for
energy harvesting using thermophotovoltaic (TPV) cells. 
We adopt a popular design scheme of metamaterial with the unit cell consisting of a 
bottom metal substrate and top plasmonic structures spaced by a dielectric layer, 
as depicted in \reffig{tpv_intro} (a). 
Tungsten (W) was selected for both the bottom metallic layer and top structural layer. 
Alumina (Al$_2$O$_3$) is used as the dielectric spacer with the thickness $\vx_1$ varying in the range 
30 nm to 130 nm. 
The dielectric spacer is  $\vx_0$ corresponds the size of a cell, 
and $\vx_2$ are the thickness of top metal layer. 
Considering the feasibility of fabricating the designed TPV, 
the cells should be single-connected and have smooth edges. 
To this end, we propose to utilize the B-Spline method to generate the shape of the cells. 
The control points of B-Spline are defined by the design parameters $\vx_3, ..., \vx_{18}$. 
To further increase the flexibility, 
we use 4 sub-cells in a single cell as depicted in \reffig{tpv_intro} (b). 
Each sub-cell is controlled by 4 control points and is reflected vertically and horizontally 
to produce a symmetrical shape. 
We set $\vx_0 \in [350, 500]$,  $\vx_2 \in [10, 80]$ and $\vx_i \in [40, \frac{\vx_0}{2}], 2 < i < 19$. 
The constraint on $\vx_i, 2 <i < 19$ is because the radius must be smaller than $\frac{1}{2}$ of the 
cell size $\vx_0$ so that the sub-cells will not overlap.
The response $\vy$ is a 500-dimensional real-valued vector.

\subsection{Structural color filter (SCF)}

\begin{figure}[h]
  \includegraphics[width=\linewidth]{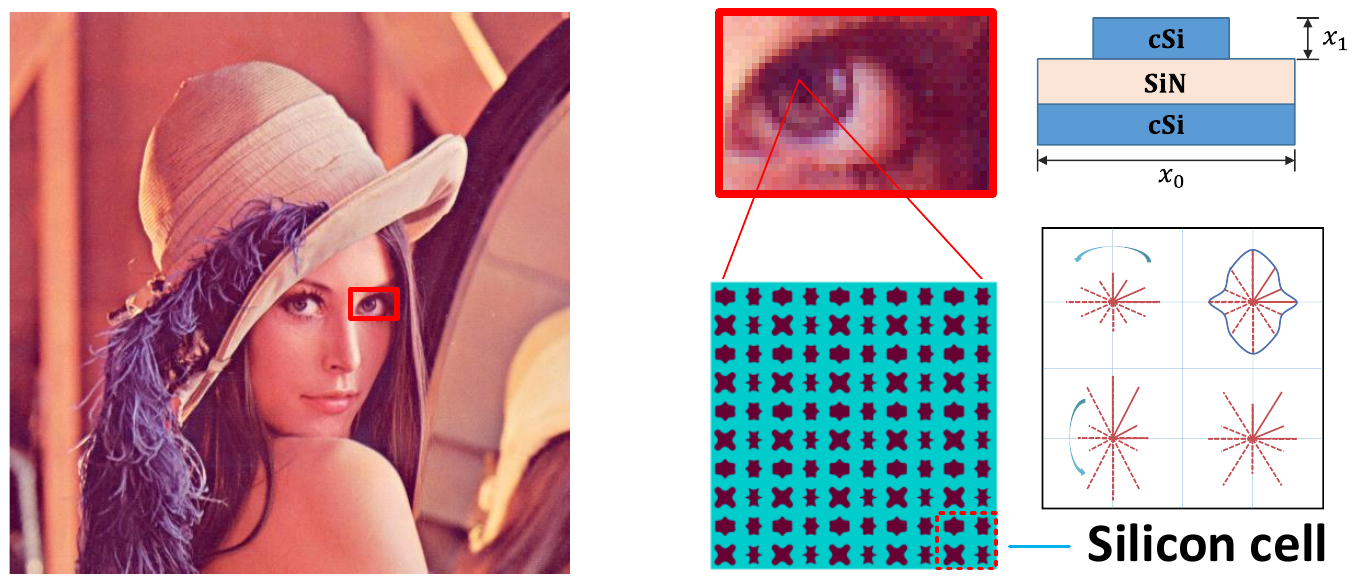}
  \caption{
    A schematic introduction of high-DPI printing with
    silicon colors.
    Each pixel is implemeted by a specific nanostructure cell,
    where the size of each cell is set to 150nm-350nm.
    \label{cf_intro}
  }
\end{figure}

Structural color filtering (SCF) using plasmonic and dielectric metamaterials has enabled 
tremendous potential for coloring applications, including passive color display, 
information storage, Bayer filtering, etc. \cite{Kumar2012natnano, Duan2017nc, Zou2022nc,Song2022natnano,color_filter_oes22,color_filter_am19} 
In this benchmark, we modeled the color filter using an arrayed silicon metasurface, 
whose supercell consists of four pillars sitting on a 70-nm SiN$_3$ thin film, 
which is supported by a silicon substrate. 
The cell size is controlled by the parameter $\vx_0$, 
and the height of the four pillars are the same and are varied as the parameter of $\vx_1$. 
The shapes of the four silicon pillars are independently generated with parameters  $\vx_2, ..., \vx_{17}$, 
which are similar to the TPV problem. 
To obtain a structural color filter, 
we need to design a color palette where each color is implemented with a specific nanostructure. 
The response $\vy$ is a 3-dimensional real-valued vector corresponding to the colors.
The inverse design problem is to find nanostructures that match with the colors that are to be printed.

\subsection{Speed of the simulator}

\begin{figure}[h]
  \includegraphics[width=\linewidth]{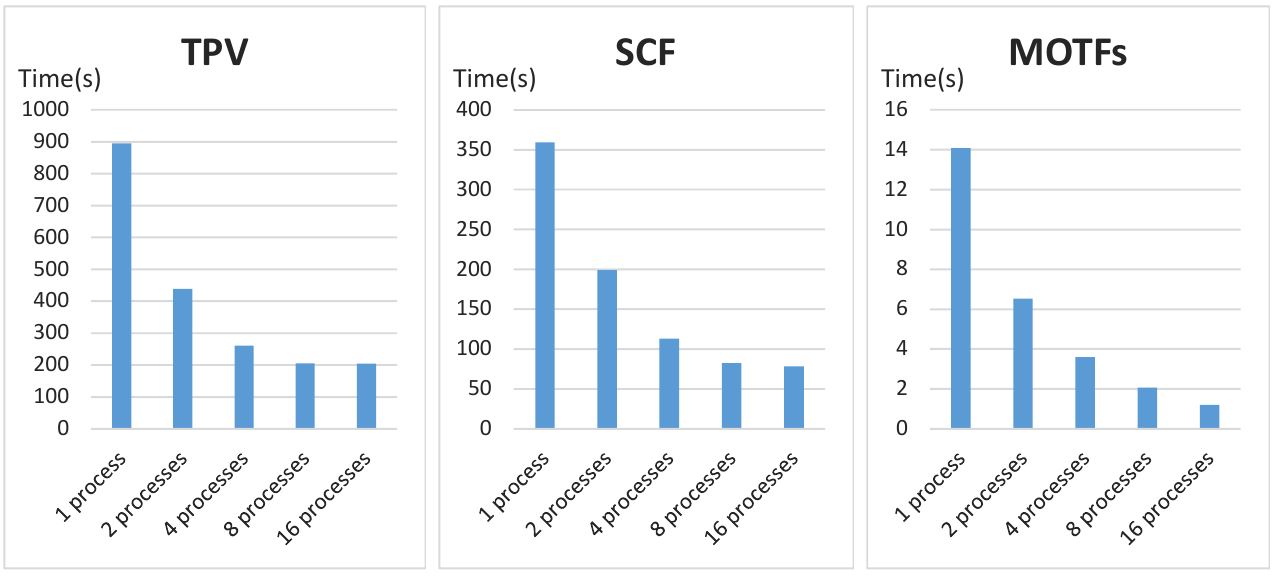}
  \caption{
    Accelerating simulation by multiprocessing
    parallelizing.
    \label{fig:parallel_speed}
  }
\end{figure}

The speed of the environment is a critical factor for a
benchmark.
On the one hand, we want the problems to be practical and
challenging;
on the other hand, we want to evaluate the designs as
fast as possible to accelerate the procedure of
developing and testing new algorithms.
We evaluate the speed of simulating 100 design parameters
with our framework, and the results are depicted in \reffig{fig:parallel_speed}.
Specifically, we use 2 Intel(R) Xeon(R) Silver 4110 CPUs,
each with 8 physical cores, and its base frequency is 2.10 GHz.
We evaluate the average time for calculating the results with
1 to 16 processes, respectively.
With 1 process, the average time of TPV,
SCF, and MOTFs are 894s, 359s, and 14s, respectively.
With 16 processes, the average time is
203s, 78s, and 1.2s, respectively.
By multiprocessing simulation,
we can speed up the calculation by 4.4, 4.6, and 11.6 times faster.
Practically, we can use more CPUs in a cluster to further
speed up calculation.
Thus, the calculation speed of the proposed problems are
sufficient for benchmark novel inverse design algorithms.
We suggest using the multilayer cooler problem to
develop algorithms with faster evaluation procedures,
then use TPV and silicon color problems for evaluation on
more challenging design problems.

\subsection{Comparing to other scientific design problems}

\textit{The performance of the nanophotonic inverse design can be accurately estimated by our simulator}. 
The performance of nanophotonic devices can be evaluated by the finite-difference time-domain (FDTD) 
method with theoretically guaranteed precision. 
This contrasts with drug and chemical designs, 
where only a heuristic score function is available\cite{drug_nips22}. 
For example, the QED function used in \cite{drug_nips22} can only predict if a chemical is "drug-like." 
The chemical found is not necessarily a real drug, let alone any performance guarantee. 
So in these problems, physical experiments are necessary to evaluate the actual design performance. 
On the contrary, 
the performance of nanophotonic devices can be accurately estimated with a dedicated FDTD simulator 
(e.g., MEEP), 
which means we can evaluate the performance of inverse design algorithms much more accurately 
without physical experiments. 
We provide more details in Appendix. 
We think such a precision guarantee makes our benchmark outstanding at reflecting the performance 
of general inverse design algorithms in the real world.

\textit{The surrogate models trained with machine learning methods may not be accurate evaluation functions}. 
Many existing benchmarks use deep neural networks to construct surrogate simulators. 
For example, MLP surrogates are used in \cite{benchmark_na}; SVM and random forest are used in DRD2, 
GSK3$\beta$, JNK3\cite{drug_nips22}. 
Such benchmarks are much easier to implement than ours: 
Generate a dataset with dedicated software, 
then train an off-the-shelf deep model (MLP, CNN, GNN, Transformer, etc.) will do the job. 
There is no need to consider the interactions and interfaces of the simulator, 
which is usually more complicated. 
However, such surrogates may not reflect the inverse design performance accurately 
(compared with the physical simulator): 
While machine learning models typically have high precision on the training and testing dataset 
generated from the same distribution, 
the precision drops significantly on the generated designs\cite{benchmark_na_j}. 
We attribute this performance drop to the fact that the designed parameters are indeed 
out-of-distribution (OOD) samples, 
which breaks the generalization guarantee of the learning-based estimators. 
Thus, our physical simulator is necessary for an accurate evaluation of nanophotonic 
inverse design methods.

\section{Implemented Algorithms}

We implement 10 different inverse design methods that cover most categories
occurred in literature.
We divide the methods into iterative optimizers and deep inverse design methods.

\subsection{Iterative optimizers}

\begin{figure}[h]
  \includegraphics[width=\linewidth]{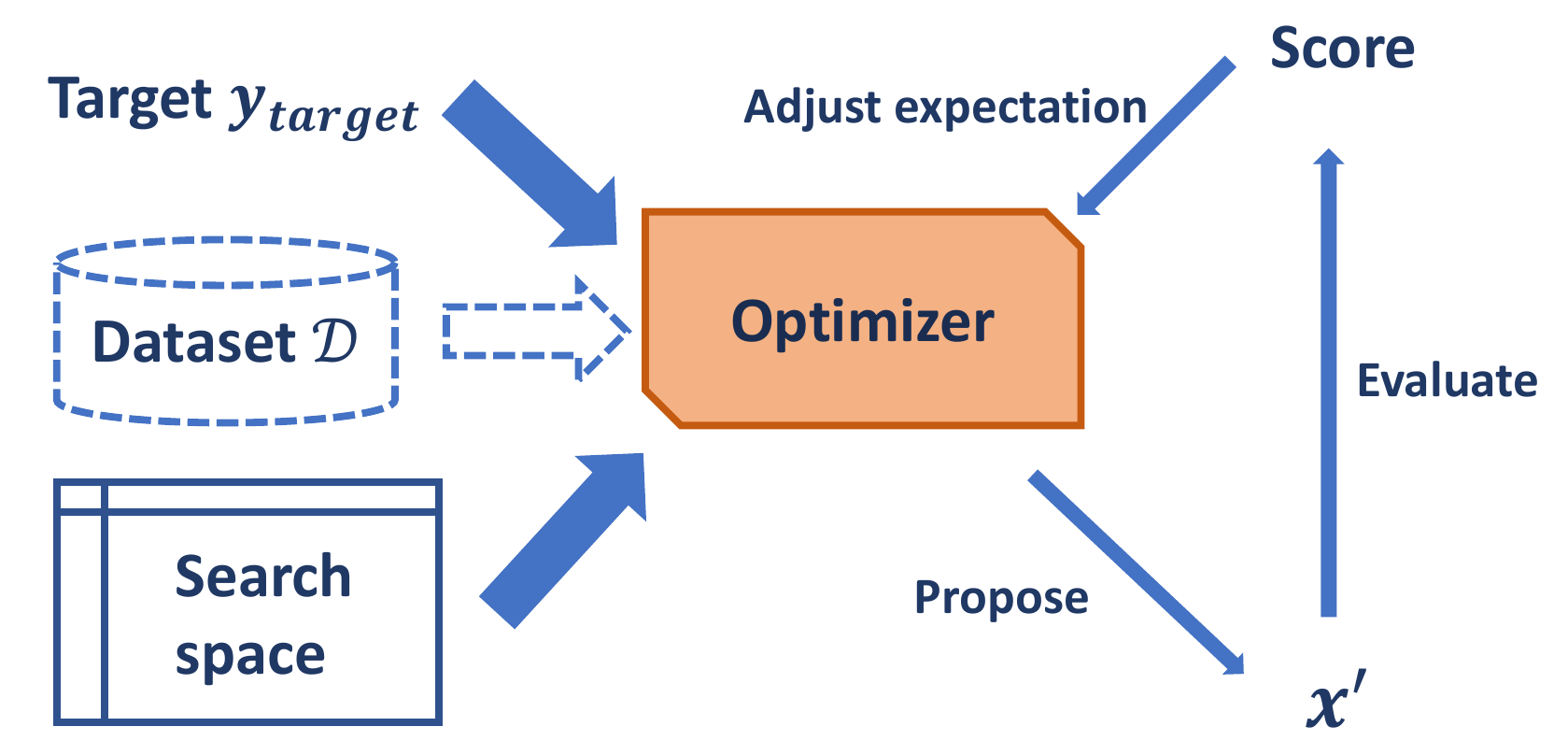}
  \caption{Iterative optimizer.\label{interactive_optimizers}}
\end{figure}

As depicted in \reffig{interactive_optimizers}, iterative optimizers work like human experts. The optimizer receives a design target, a specification of search space, and a dataset containing some promising design is optional. Based on the current belief of the search space, the optimizer produces several design parameters $\vx^\prime$ that are considered promising. The design parameters $\vx^\prime$ will be evaluated and produce a score, which indicates the performance of this design parameter. The optimizer needs to interact with the simulator to obtain a performance score $\mathcal{S}(\vx^\prime)$. The design parameter $\vx^\prime$ and the score are fed to the optimizer to adjust the belief of the optimizer. Then, the optimizer generates new design parameters that are considered promising. The whole procedure iterates until satisfactory results are obtained. The iterative optimizers replace expert efforts in analyzing and producing new design parameters. The design procedure can be fully automated.

The experts may also suggest some promising design parameters to the optimizer. Therefore, these methods are highly flexible and can be used to aid manual design. There are also some downsides to iterative optimizers. The iterative optimizer can only design a design target at one time, which will be too time-consuming for inverse design problems with many design targets, such as SCF. The training time and inference time of machine-learning-based optimizers such as Bayesian optimizer scales as $O(n^3)$ where $n$ is the number of explored data points. Some iterative optimization methods are designed to support discrete design parameters, such as TPE, SRACOS, and ES. This is an important advantage since most of the inverse design problems are involved with some discrete variables such as material type or layer number.

We implement the following iterative optimization methods:
\begin{itemize}
  \item Random search (\textbf{RS}):
        The simple but widely used hyper-parameter search method
        which randomly selects design parameters within the parameter space.
  \item Sequential randomized coordinate shrinking (\textbf{SRACOS})\cite{exp_sracos}:
        The SRACOS algorithm trains a classifier to classify good parameters with
        larger scores and bad parameters with smaller scores.
        Then, the algorithm draws samples from the parameter space uniformly
        and select good parameters with the trained classifier as its
        proposal.
        The evaluated parameters are inserted into the training dataset
        to improve the classifier, and new proposals are generated.
        We implement SRACOS with ZoOpt\cite{exp_zoopt}.

  \item Bayesian Optimization (\textbf{BO})\cite{exp_bayesian_optimization,bo_id_2023}:
        The Bayesian optimization method trains a model with a Bayesian method,
        typically a Gaussian process, which could learn the probability distribution
        of the score $p(\mathcal{S(\vx)}|\vx)$.
        With the probability information, BO algorithms can evaluate
        how likely a parameter may improve the score.
        Then, promising parameters are selected to be evaluated.
        We implement BO with the BayesOpt library \cite{exp_bayesopt}.

  \item Tree-structured Parzen Estimator Approach (\textbf{TPE}) \cite{exp_TPE,bo_tree_id_2021}:
        Instead of modeling $p(\mathcal{S(\vx)}|\vx)$ directly, TPE
        models $p(\vx|\mathcal{S(\vx)})$ and $p(\mathcal{S(\vx)})$ with
        a tree-structured estimator.
        We implement the TPE method with Hyperopt\cite{exp_hyperopt,exp_hyperopt2}.

  \item Evolution Strategy (\textbf{ES})\cite{exp_oneplusone,es_id_2022}:
        Evolution strategies are a set of algorithms that gradually modify
        current solutions with heuristic methods.
        We implement a representative ES algorithm called the one-plus-one
        (1+1) algorithm with implementation from Nevergrad\cite{exp_nevergrad}.
\end{itemize}

\subsection{Deep inverse design methods}

\begin{figure}[h]
  \includegraphics[width=0.7\linewidth]{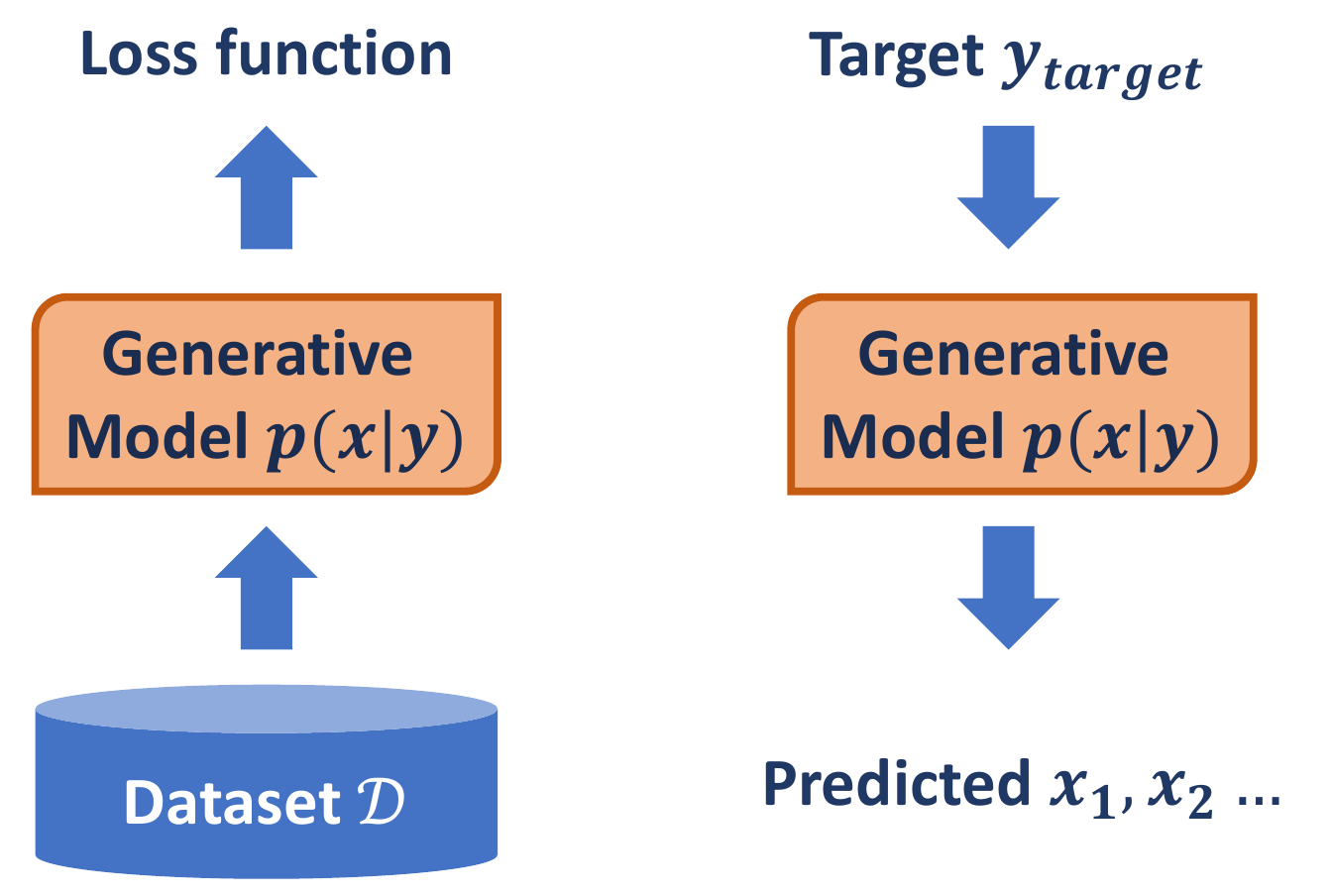}
  \caption{Deep inverse design method.\label{deep_id_methods}}
\end{figure}

With the prosperity of deep learning, deep-learning-based methods are
adopted to inverse design problems\cite{inverse_design_in_nanophotonics},
such as VAE\cite{id_vae}, GAN\cite{id_gan}, AAE\cite{id_aae}, etc.
The typical procedure of deep learning-based inverse design methods is
depicted in \reffig{deep_id_methods}.
The deep learning methods are trained to minimize specific loss functions.
The trained model can predict design parameters given a design target.
Some deep learning methods can produce multiple design parameters $\vx_1, \vx_2,..., \vx_n$ from $p(\vx|\vy_{target})$,
which is more reasonable since the best design parameter may not be unique.

There are two outstanding advantages of deep learning methods.
First, the inference of deep-learning-based methods is typically much faster than iterative optimizers,
which enables efficient design of many different design targets.
The design for different targets can be parallelized to improve speed further.
Secondly, since the prosperity of deep learning methods,
the accuracy of neural networks significantly outperforms traditional
learning methods on various applications. Deep-learning-based methods can
benefit from the progress of neural network architectures directly and
lead to better performance.

There are also some disadvantages.
First, Deep learning methods have large amounts of parameters,
which requires a large amount of data to train.
However, in inverse design problems, generating a dataset is typically time-consuming.
Thus, the amount of data is a critical bottleneck in the performance of deep learning methods.
Secondly, the training time for deep learning methods is long,
so it is not easy to add new training data during design like iterative optimization methods,
which lacks the ability to improve the model.
Thirdly, deep learning methods only work for continuous variables currently,
but does not well suit discrete variables.

\begin{itemize}
  \item Inverse Model (\textbf{IM})\cite{exp_tandem}:
        The most straightforward method is to train an inverse model that
        predicts design parameters given the design target.
        There is an obvious drawback to this method:
        there may be multiple different design parameters that
        lead to the same or similar response.
        This method will fail to learn the correct design parameter
        if the solution is not unique.

  \item Gradient Descent (\textbf{GD})\cite{benchmark_na,benchmark_na_j}:
        We can train a forward model $f(x)=\mathcal{F}(x)$
        to replace the simulator $\mathcal{F}$.
        Then, we can use this substitute forward model to find a good solution.
        If the forward model is differentiable, and the derivatives with
        respect to the input design parameters are easy to solve,
        we can apply gradient descent to maximize the performance score
        directly.
        In this GD method, we train an MLP as the forward model.

  \item \textbf{Tandem}\cite{exp_tandem}:
        Tandem has a forward model and a prediction model.
        The forward model predicts the response $\vy=f(\vx)$ given the design
        parameters $\vx$ as in GD.
        Different from GD, Tandem proposes a prediction model $\vx^\prime=g(\vy)$ to
        predict the design parameters directly.
        The prediction model is trained with cycle loss to make sure that
        $f(g(\vy))\approx \vy$ while keep the forward model $f$ fixed.

  \item Conditional Generative Adversarial Networks (\textbf{CGAN})\cite{exp_cgan1,exp_cgan2,gan_id_2020,gan_id_2022}:
        Generative adversarial networks (GAN) have a generator network
        and a discriminator network. The discriminator network discriminates between real data and generated data.
        The generator is trained to cheat the discriminator.
        The two networks are trained alternatively.
        Theoretically, a GAN will converge to the data distribution $p(x)$.
        Here we use conditional GAN (CGAN) to learn the conditional
        distribution $p(\vx|\vy)$.
  \item Conditional Variational Auto-Encoder (\textbf{CVAE}) \cite{exp_cvae,cvae_id_2022}:
        Variational Auto-Encoders (VAE) have an encoder network that models
        $p(z|x)$ and a decoder network which models $p(x|z)$,
        where $z$ is a latent variable that has a simple prior distribution,
        typically isotropic Gaussian $\mathcal{N}(0, I)$.
        The VAE maximizes an evidence lower-bound (ELBO) of
        the log-likelihood $\sum \log p(x)$ during training.
        After training, the decoder can be used to sample design parameters
        from $p(x)$ with $z \sim \mathcal{N}(0, I)$ and $x \sim p(x|z)$.
        Here we adopt conditional variational auto-encoder (CVAE) to
        train a model of the conditional distribution $p(\vx|\vy)$.
\end{itemize}

All the methods are implemented using Pytorch\cite{pytorch} within the ray framework\cite{exp_ray_tune}.
Because of the page limit, more details and codes are provided in the Appendix.

\section{Experiments\label{sec:experiments}}

\subsection{Dataset}

We generate datasets with uniform sampling from the design parameter spaces.
Specifically, we sample from $\mathcal{U}(x_{min}, x_{max})$ for a
continuous design parameter $x$ with maximum value $x_{max}$ and minimum value $x_{min}$.
For a discrete variable $x \in \{c_0, ..., c_{k-1}\}$ where $c_i$ denotes a categorical value,
e.g., material name in the multilayer film problem,
the value of $x$ is chosen uniformly from all the $k$ different
values with probability $\frac{1}{k}$.
The statistics of the three datasets are summarized in \reftab{dataset_stats}.

\begin{table}[htbp]
  \centering
  \caption{Statistics of datasets\label{dataset_stats}}
  \begin{tabular}{lrrr}
    \toprule
          & \multicolumn{1}{l}{Dataset size} & \multicolumn{1}{l}{\# Input} & \multicolumn{1}{l}{\# Output} \\
    \midrule
    MOTFs & 60700                            & 20 (80)                      & 2001                          \\
    TPV   & 69600                            & 19                           & 500                           \\
    SCF   & 25500                            & 18                           & 3                             \\
    \bottomrule
  \end{tabular}%
  \label{tab:addlabel}%
\end{table}%

The input of the multilayer film problem consists of 10 categorical parameters (material)
and 10 numerical parameters (layer thickness).
Some methods such as TPE and ES have direct support for categorical variables,
so we use the 20 parameters as inputs directly.
For forward prediction tasks, the categorical variables are transformed to their
one-hot encodings, so we have $7*10+10=80$ inputs since there are 7 different materials
the multilayer film problem.

\begin{figure*}
  \includegraphics[width=\textwidth]{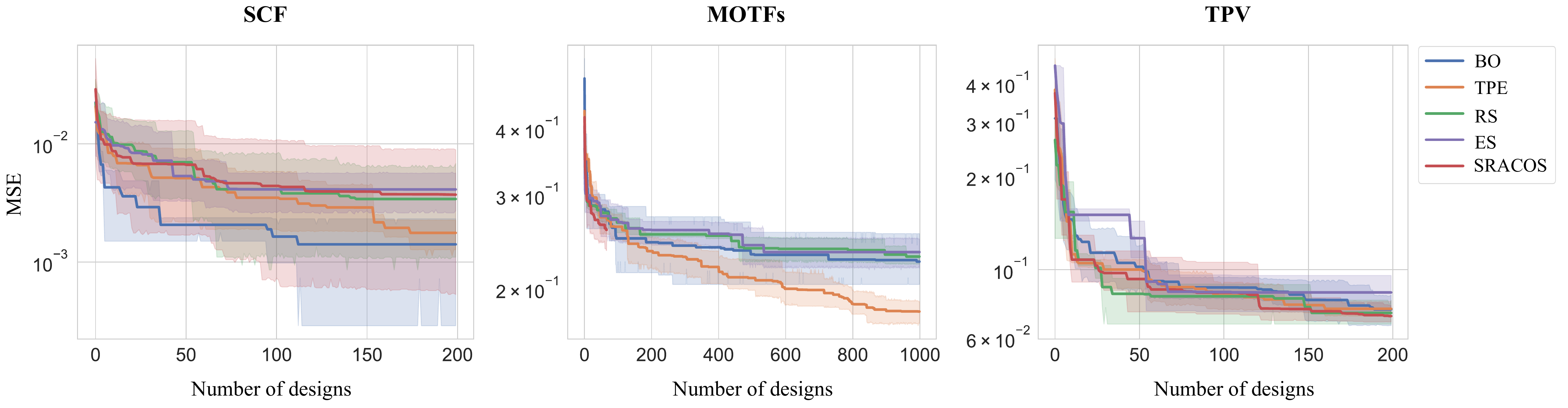}
  \caption{
    Benchmark on the inverse design tasks without training data.
    \label{fig:benchmark_real_0_train}
  }
\end{figure*}

\begin{figure*}
  \includegraphics[width=\textwidth]{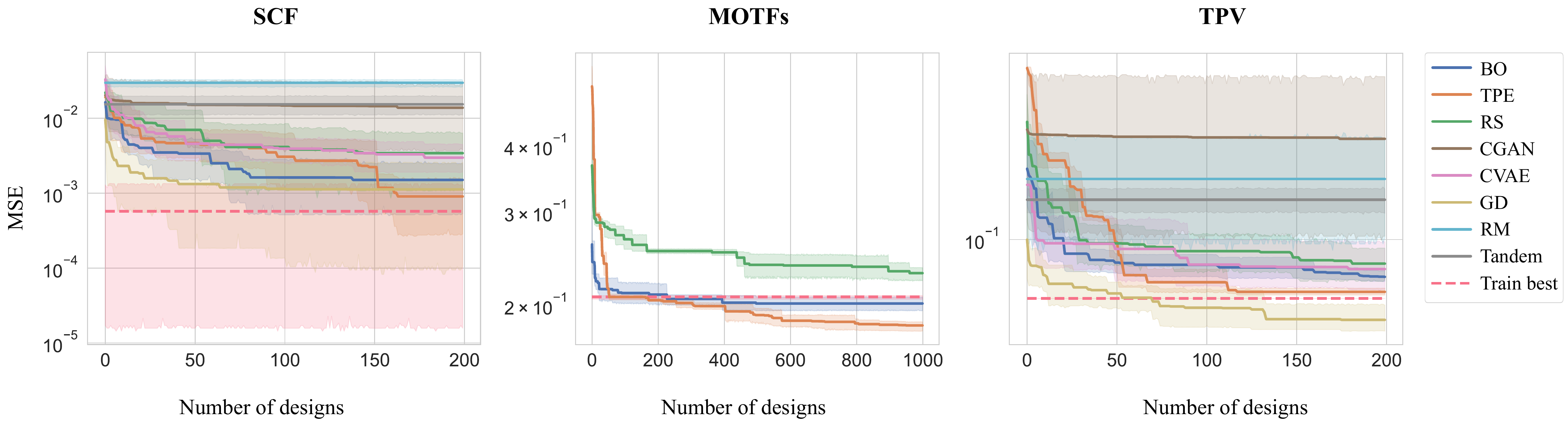}
  \caption{
    Benchmark the inverse design tasks using all the training data.
    \label{fig:benchmark_real_all_train}
  }
\end{figure*}

We split 90\% as the training dataset and 10\% as the testing dataset.
The testing dataset serves as testing data for forward prediction tasks and
inverse design tasks with IID (independent and identically distributed) targets,
as described in \refsec{sec:exp_iid} in detail.
We further split 90\% from the training dataset to train the deep inverse design methods and the forward prediction methods; the left 10\% is used for validation.
The interactive optimization methods generally do not have a training stage,
but they may accept a small collection of suggested design parameters to be tried
at first. So we select the 100 best design parameters with the highest scores from the
training dataset as the suggested design parameters.
It's also possible to pass the whole dataset to the optimizers.
However, the speed is extremely slow since the optimization methods are not
designed for large-scale training, and the computational complexity may
also be an obstacle for, e.g., Bayesian optimization, which has $O(N^3)$ time complexity
where $N$ is the number of processed data points.

\subsection{Inverse Design with Real Targets}

All three inverse design problems have pre-defined design targets,
which align with their real-world applications.
The goal of inverse design is to find design parameters that can
produce a response that is as close to the design target as possible.
In our experiments, we use the mean squared error (MSE) as the performance measure.
Specifically, the performance of a design parameter $\vx$ is calculated as follows:
\begin{equation}
  \mathcal{L}(\mathcal{F}(\vx), \vy_{target}) = \sum_j^{m} ({\vy_{target}}_j - \mathcal{F}(\vx)_j)^2
\end{equation}
In the following inverse design benchmarks,
we set the number of designs that can be tried by each algorithm to
200, 200, and 1000 in TPV, SCF, and MOTFs, respectively.
All the algorithms are run 5 times with different random seeds,
we plot the mean value and its 95\% confidence interval.

\subsubsection{Zero training}

The ideal work procedure of an inverse design method is to take the
design target as input, and optimize the MSE in a fully automated way without
any initial training data.
We call this the zero training benchmark.
All the iterative optimization methods are designed to work in zero training cases;
all the deep learning methods can not work without training data.
So we only compare iterative optimization methods in this benchmark.
The results are depicted in \reffig{fig:benchmark_real_0_train}.

We have the following conclusions
\begin{enumerate}
  \item The TPE and BO algorithms perform similarly well on the SCF problem,
        outperforming other algorithms by a large gap.
        The target space of the silicon color problem
        is simpler (the dimension of $\vy$ is 3) than the other two problems,
        which enables the probability estimation to be more accurate.
        Thus, the search algorithms based on probability work better.
  \item The TPE algorithm performs best on the MOTFs problem,
        while other algorithms perform similarly.
        There are discrete variables in the MOTFs problem,
        which can be modeled better with tree-structured methods such as TPE.
  \item In the TPV problem, all the algorithms perform similarly,
        as well as the random search method.
        The TPV problem is intrinsically more complex than the other two problems
        since all the algorithms cannot find search directions better than
        random search within 200 trials.
\end{enumerate}

\subsubsection{Full train}

\begin{figure*}
  \includegraphics[width=\textwidth]{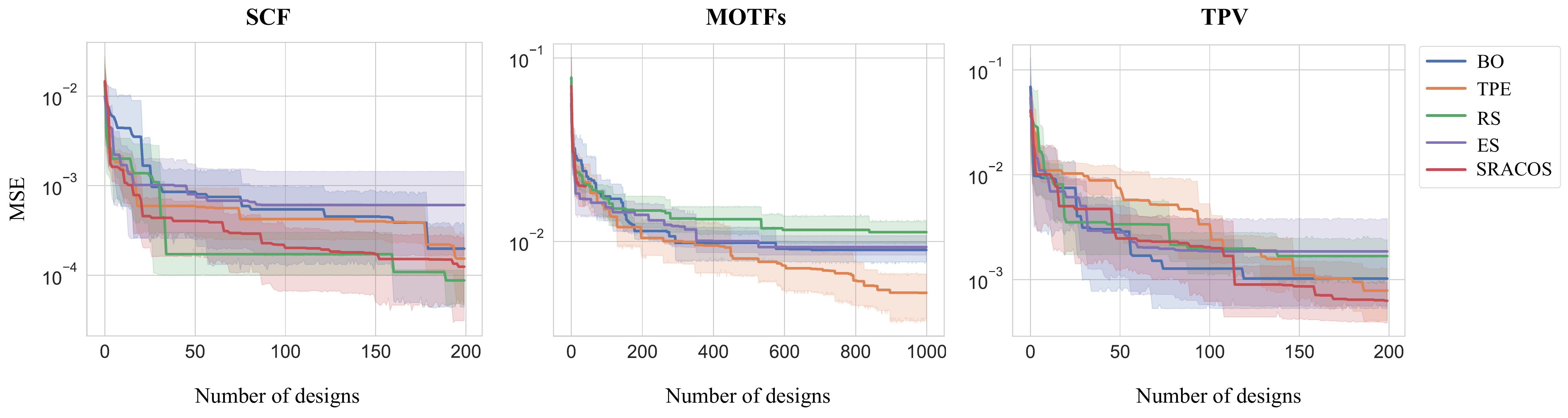}
  \caption{
    Benchmark on the inverse design tasks with IID-generated targets.
    The training dataset is not used.
    \label{fig:benchmark_iid_0_train}
  }
\end{figure*}

\begin{figure*}
  \includegraphics[width=\textwidth]{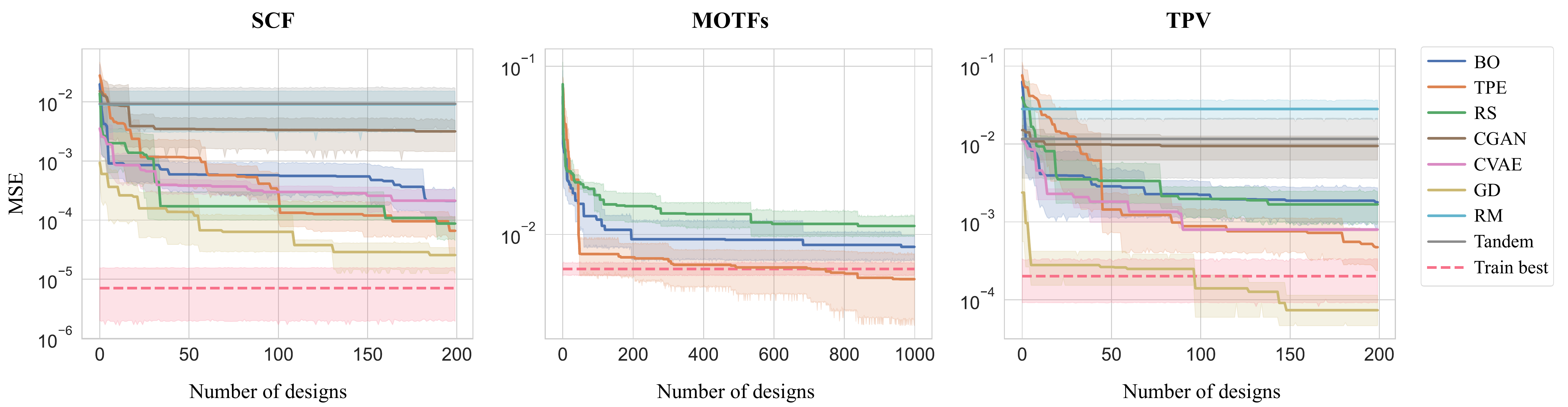}
  \caption{
    Benchmark on the inverse design tasks with IID-generated targets.
    All the training dataset is used.
    \label{fig:benchmark_iid_all_train}
  }
\end{figure*}

Typically, generating a dataset with random sampling requires considerable
computational resources.
In this benchmark, we compare the performance of inverse design methods that
are trained on the whole training dataset.
The results are depicted in \reffig{fig:benchmark_real_all_train}.
We also plot the performance of the best design parameter within
the training dataset (denoted as \textit{Train best} in dashed line),
which also tends to be a reasonably strong baseline.

We have the following observations and conclusions:
\begin{enumerate}
  \item  In our benchmark results, all of the algorithms struggle to make
        further progress beyond the best training design within limited trials.
        None of the algorithms exceed the train-best in SCF;
        TPE is the only method that makes significant progress in MOTFs;
        GD is the only method that exceeds the train-best in TPV.
  \item The train-best design of silicon color is good enough (MSE < $10^{-3}$),
        which makes it hard to make further improvements.
        How to make further progress upon high-precision baselines
        is a good problem for future research.
  \item The TPE method can learn meaningful information from
        training data and produce promising design parameters for
        optimization with discrete variables.
  \item Although TPV is a hard problem,
        GD can find better design parameters within 200 trials,
        which indicates that deep models can learn better
        searching directions of the problem using sufficient training data.
\end{enumerate}

\subsection{Inverse Design with IID Targets\label{sec:exp_iid}}

The inverse design with real targets is a real-world problem,
but may not be an ideal benchmark problem.
There are several reasons:
First, the perfect design parameter $\vx^*$ with $\mathcal{F}(\vx^*)=\vy_{\text{target}}$
may not exist.
Indeed, a perfect design with optimal performance does not exist
in most real-world problems,
which is simply because of the limitation of physical constraints.
So we cannot evaluate how close we are to the best possible solution.
Secondly, the performance of the inverse design on the IID targets may serve as an upper bound for
the performance on the real target.
Indeed, the real targets are out-of-distribution (OOD) data,
which are much harder than the IID-generated (in-distribution) targets.
Thirdly, the IID assumption is the cornerstone of most machine learning methods
and may provide better theoretical properties for further analysis.
Thus, the performance of the inverse design method with IID targets is
of particular interest for developing and evaluating inverse design algorithms.
We sample 5 random design parameters $\vx$,
then calculate their responses $\vy$.
These responses are used as IID design targets,
which are guaranteed to be realizable in our design parameter space.

The results without training data are depicted in \reffig{fig:benchmark_iid_0_train},
and the results using all the training data are depicted in \reffig{fig:benchmark_iid_all_train}.
We have the following observations and conclusions:
\begin{enumerate}
  \item Design for the IID targets is a much easier task since it can intuitively lead to much lower MSE.
  \item The performance on the SCF problem is
        very different from the case with real targets.
        Thus, the evaluation with IID targets cannot
        substitute the evaluation with real targets in the SCF problem.
  \item The conclusion on the MOTFs problem is
        similar to results with real targets: TPE is the best-performing
        method. The conclusion on the TPV problem is also similar
        to evaluating with the real targets that
        GD is the best-performing method.
        This indicates that we may replace real targets with IID targets
        on specific problems.
\end{enumerate}

\section{Related Work}

\citet{idbenchmarks_iclr19} and \citet{idbenchmarks_icml21}
propose several simple physical problems to
evaluate the performance of inverse probability density estimation.
However, the physical problems are too simple compared to real-world
inverse design problems.

Almost all the research in traditional photics and material science
are not open-source\cite{baumert1997femtosecond,LalauKeraly2013,sell2017large,xiao2016diffractive,Hughes2018,Mansouree2021,Campbell2019}.
\citet{benchmark_no_opensource} proposes
a toolkit for inverse design problems that relies on MATLAB and Lumerical.
A prominent difficulty to develop
open-source benchmarks on inverse design problems lies in that
the modeling is typically implemented using expensive commercial
software such as CST, COMSOL, MATLAB, and Lumerical

To side-step such difficulty, several benchmarks have been proposed to substitute simulators with machine learning models trained on a randomly generated dataset\cite{benchmark_na, benchmark_na2, idbenchmarks_ridnoise}. However, there is a significant gap between the predicted performance by the surrogate model and the accurate simulators.\cite{benchmark_na_j} Surrogate models are generally trained and validated on independent and identically distributed (IID) data, while the inverse-designed parameters are typically non-IID.
This discrepancy could lead to misleading benchmark conclusions. To the best of our knowledge, no one has provided an open-source benchmark for inverse design problems with direct access to an accurate simulator.
And our benchmark is also fast, reproducible, scalable, and extensible.

\section{Future Directions}

\subsection{Combination of deep learning and interactive optimization}

As our comparison at \refsec{sec:experiments}, 
the deep-learning-based methods have better performance on big data and continuous variables,
and the iterative optimization methods can tackle discrete variables and start
working without training data.
We suggest dimension reduction and learning to optimize to combine their advantages.

\noindent \textbf{Dimension reduction.}
It is known that black-box optimizer will be less effective on high-dimensional optimization\cite{high_dim_bo}.
The reason is that the black-box optimization method has little knowledge of the underline structure of
the problem, which makes the search inefficient on high-dimensional problems.
If we can reduce the dimension of the original problem to a low-dimension space\cite{manifold},
the performance of the interactive optimization methods may be improved.
A similar idea has been explored by \citet{high_dim_bo},
where a multilayer network with sigmoid activation is used to reduce dimension.
We believe there is still room for further improvements with the development
of dimension reduction methods in recent years.

\noindent \textbf{Learning to optimize.}
The interactive optimization problem can be seen as a decision-making problem:
at each step, the algorithm needs to decide what is the next design parameter to be evaluated
based on previous observations.
After receiving the score, the optimizer needs to adjust the next design parameters to
achieve better performance.
Such a decision-making procedure is similar to reinforcement learning,
where the policy network should learn to make optimal actions facing different
environment states\cite{rl_review}.
There are also some related topics on learning to optimize neural networks\cite{learning_to_optimize}.
However, these learning-based methods require gradients of the objective,
which is not available in inverse design problems.
Because of its flexibility and data efficiency,
learning-based optimization is another promising direction of inverse design algorithms.

\subsection{Better forward model and search algorithms}

\noindent \textbf{Developing better forward model.} One of the reasons for the success of deep learning can be attributed to
the ability to learn high-level representation from complex structured data.
For example, recent deep learning methods can learn from graph data
(there are relations between data points)\cite{gnn_review},
multi-modal data (e.g., data consists of images, videos, texts, etc.)\cite{multi_modal_review},
time series (the concerned value varies along with time)\cite{time_series_review}, etc.
Such complex data structures also exist in inverse design problems, 
and the prediction performances vary by methods and problems.
Thus, adopting cutting-edge deep learning methods will lead to better performance
of the forward model and enables working with more complex problems.

\noindent \textbf{Search algorithms on substitute model.}
A straightforward inverse design method is to train a
forward model, then use this substitute model to accelerate the design procedure.
The forward model can be used to evaluate a design parameter's performance.
We can also utilize the information (e.g., gradients with respect to design parameters) learned
by the model to accelerate the optimization of the design parameters.
However, as the inverse design problems are typically non-convex,
thus, gradient descent is not guaranteed to converge to global minima.
And for discrete variables or graph-structured variables,
the gradient information is not available.
Thus, better search algorithms for efficient optimization of the forward model
is necessary for the efficient inverse design.

\subsection{Better work with human prior and human in loop}

How to incorporate human knowledge has become an important topic in machine learning
with the name "human-in-the-loop"\cite{human_in_loop},
which aligns with the needs of inverse design problems.

We suggest that how to utilize human knowledge is critical for efficient
inverse design.
Since the simulation and experiments are typically expensive,
a large amount of data may not be available for a specific design problem.
And the experience from past design problems is hard to be transferred to the current
design problem with learning-based methods because of the change in parameter space and target space.
Thus, utilizing the professional knowledge from human experts is
an essential direction in the future development of inverse design methods.

\section{ACKNOWLEDGEMENTS}

This research was supported by National Key R\&D Program of China (2022ZD0114805,2022YFF0712100), 
National Natural Science Foundation of China (62275118,62276131), 
the Fundamental Research Funds for the Central Universities,
Young Elite Scientists Sponsorship Program by CAST, 
the Fundamental Research Funds for the Central Universities (NO.NJ2022028, No.30922010317). 
Professor Yang Yang and Professor De-Chuan Zhan are the corresponding authors.

\bibliographystyle{ACM-Reference-Format}
\balance
\bibliography{ref}

\end{document}